# Research on Dangerous Flight Weather Prediction based on Machine Learning

Haoxing Liu[1], Renjie Xie[1], Haoshen Qin[2] and Yizhou Li[3*]

[1]Flight Department, Shanghai Jixiang Airlines Co., Ltd., Shanghai, China;
[2]Department of Computer & Information, Science & Engineering, Herbert Wertheim College of Engineering, University of Florida, Gainesville, 32611, USA;
[3]Department of Electrical, Computer, and Systems Engineering, Case Western Reserve University, Cleveland, 44106, USA.
*Corresponding author's e-mail address: yxl3527@case.edu

**Abstract.** With the continuous expansion of the scale of air transport, the demand for aviation meteorological support also continues to grow. The impact of hazardous weather on flight safety is critical. How to effectively use meteorological data to improve the early warning capability of flight dangerous weather and ensure the safe flight of aircraft is the primary task of aviation meteorological services. In this work, support vector machine (SVM) models are used to predict hazardous flight weather, especially for meteorological conditions with high uncertainty such as storms and turbulence. SVM is a supervised learning method that distinguishes between different classes of data by finding optimal decision boundaries in a high-dimensional space. In order to meet the needs of this study, we chose the radial basis function (RBF) as the kernel function, which helps to deal with nonlinear problems and enables the model to better capture complex meteorological data structures. During the model training phase, we used historical meteorological observations from multiple weather stations, including temperature, humidity, wind speed, wind direction, and other meteorological indicators closely related to flight safety. From this data, the SVM model learns how to distinguish between normal and dangerous flight weather conditions.

## 1. Introduction

There are many meteorological factors that affect flight safety, including air pressure, temperature, wind, cloud cover and visibility. During the take-off and landing phases of the aircraft, it is susceptible to weather phenomena such as downbursts, low-altitude windshear, fog, smoke, frost, snowfall, blowing snow, sandstorms, blowing sand and low clouds; During the climbing, cruising and descent phases, weather phenomena such as bumps, local circulation, topographic waves, jet streams, thunderstorms, squall lines, typhoons, ice, sleet, sandstorms and floating dust are common. Even if aircraft remain on standby on the airport tarmac, dangerous weather such as thunderstorms, hailstorms, tornadoes and typhoons can seriously affect airport safety [1].

In the modern air transport industry, flight safety is always one of the most critical considerations. As the impact of global climate change intensifies, the frequency of extreme weather events has become one of the major challenges facing air transportation. Dangerous flight weather, such as severe storms, thunderstorms, hail, low visibility and strong turbulence, can significantly affect the safety and efficiency of flights [2]. Therefore, accurate prediction of these weather phenomena is key to improving aviation safety and operational efficiency.

Strong atmospheric currents are an important factor affecting aviation flight safety. End-flow is a form of intense movement in the atmosphere and refers to weather phenomena with large variances in the velocity of meteorological target particles in the atmosphere in a specific area. The velocity distribution of the particles contained in the target is an indicator to judge the strength of the end flow, and the wider the velocity distribution, the greater the fluctuation of the airflow in the entire target area, so the end flow is also called atmospheric turbulence [3]. The following two conditions need to be met for the air mass to produce strong atmospheric turbulence including thermodynamics and kinetics. The thermodynamics air layer must be unstable, especially when the atmospheric humidity increases with the decrease of altitude, which is particularly easy to cause vertical convection of the atmosphere, and then produce end flow [4]. Kinetics needs to be significant wind speed shear in the air mass, and strong end currents are often accompanied by thunderstorms, with the strongest end flow in the middle of the vertical profile of the thunderstorm region, and turbulence may also be present at the outer edge of the thunderstorm.

The main types of turbulence affecting aviation flight include aircraft wake, thermal turbulence and dynamic turbulence. Atmospheric turbulence can be further divided into clear-sky turbulence and wet-air turbulence according to whether the turbulent region contains precipitation and raindrops [5]. When flying in a turbulent area of clear sky, the aircraft is likely to experience turbulence. Clear sky bumps usually occur at altitudes of 6,000 meters and above, mostly in cloudless or cirrus clouds, and their thickness is usually between 200 and 1,500 meters, while in the troposphere most do not exceed 1,000 meters, and are common in the lower troposphere.

In the middle troposphere, the frequency of turbulence is the lowest; In the upper troposphere, the closer you get to the tropopause or the height at which the maximum wind speed is located, the more frequent you will encounter strong turbulence [6]. Moderate intensity and strong turbulence mostly occur near the jet axis: (1) the cold advection zone in the upper-air trough near the jet axis; (2) The confluence of two rapids. When the two jet streams are close to within 500 km, there are often clear sky bumps near the confluence area; (3) Cut off low pressure at high altitude. When the high altitude cuts off the low pressure and causes the jet stream to be reversed, moderate or above turbulence may occur; (4) High-altitude ridges. When the jet stream is steered along the upper-level ridge, moderate to strong turbulence may occur at the ridge top.

In recent years, machine learning technology has shown its powerful data processing and forecasting capabilities in many fields, and aviation weather prediction is no exception. Using machine learning methods, researchers are able to learn from large amounts of historical meteorological data and discover complex patterns and associations to improve the accuracy of their predictions [7]. These methods include, but are not limited to, decision trees, neural networks, support vector machines, etc., which can efficiently process and analyze large-scale meteorological datasets to predict impending hazardous weather conditions.

## 2. Related Work

Campbell et al. [8] developed the WX1 system, which uses an artificial intelligence approach to automatically identify wind shear. At the end of the 20th century, NASA and a number of scientific research institutions conducted a large number of experiments on low-altitude windshear and turbulence detection, and conducted in-depth analysis and research on the characteristics of low-altitude windshear and turbulence. The research team proposed several targeted radar signal processing methods, including adaptive filtering algorithm, fast Fourier transform algorithm (FFT), pulse pair algorithm (PPP) [9], and notch method. These methods perform signal processing on the echo data, filter the clutter, and obtain the characteristics of windshear and turbulence, output the regional coordinates and grades of windshear and turbulence according to the relevant airworthiness standards, and issue corresponding warnings information.

Although Doppler lidar is highly accurate, its detection range is relatively short in the field of aviation meteorology, especially in the presence of rain, fog or low clouds. Therefore, K. Satheesan and B.V. Krishna Murthy [10] used Doppler wind profile radar to observe the troposphere and stratospheric bottom. In recent years, the National Center for Atmospheric Research has also developed a turbulence detection algorithm for the WSR-88D radar. The algorithm utilizes spectral width, radial velocity, and reflectance data, and undergoes rigorous quality control to provide an assessment of the confidence of each measurement point, resulting in detailed information on the turbulent dissipation rate and its trustworthiness.

Additionally, a comprehensive study highlighted the potential of machine learning in transforming weather forecasting. The research pointed out that machine learning models developed with high-quality reanalysis datasets could offer forecasts that are not only cost-effective but also highly competitive in accuracy compared to traditional Numerical Weather Prediction (NWP) systems [11]. These models are particularly evaluated for their efficiency in predicting extreme weather events, although they face challenges like increasing bias with forecast lead time and poor performance in predicting the intensity of tropical cyclones.

## 3. Methodologies

*3.1. Notions*

Above all, we present the main parameters used and their corresponding explanations in the following Table 1.

**Table 1.** Primary Notions

| Parameter Symbols | Explanations |
|---|---|
| $W$ | weight vector of the hyperplane |
| $x_i$ | Feature vectors |
| $y_i$ | label associated with each data point |
| $b$ | Bias term |
| $\gamma$ | width of the kernel |
| $\lVert \cdot \rVert^2$ | squared Euclidean distance |

*3.2. Support Vector Machine*

Support vector machine is an effective classification technique that separates two types of data by constructing a hyperplane with maximum spacing. Mathematically, Support vector machine model seek to solve the following optimization problems, which express as Equation 1.

$$min_{w,b} \frac{1}{2}\lVert W \rVert^2 \; subject\; to \; y_i(W \cdot x_i + b) \geq 1, \forall i \qquad (1)$$

Here, the objective is to minimize $\frac{1}{2}\lVert W \rVert^2$, where $W$ represents the weight vector of the hyperplane. Minimizing this term effectively maximizes the margin between the classes. These constraints ensure that all data points $x_i$ are classified correctly. Parameter $y_i$ is the label associated with each data point, which can be either +1 or -1. The constraint stipulates that each data point must lie on the correct side of the margin.

Additionally, the weight vector $W$ is normal to the hyperplane and determines its orientation in the feature space. Bias term $b$ adjusts the distance of the hyperplane from the origin without depending on the input data, effectively shifting the hyperplane closer to or further from the origin to achieve an optimal separation. Feature vectors $x_i$ are the individual data points in the dataset, represented in the feature space. Labels $y_i$ indicate the class of each feature vector, helping to guide the placement of the hyperplane by the constraints.

*3.3. Radial Basis Function*
The radial basis function (RBF) kernel is a popular choice in support vector machine models for handling non-linearly separable data. The radial basis function kernel is effective because it transforms the data into a higher dimensional space where the linear separator is possible, even if the data cannot be linearly separated in the original space. The calculation process in proposed model is expressed as Equation 2.

$$K(x_i, x_j) = \exp(-\gamma ||x_i - x_j||^2) \qquad (2)$$

Note that $x_i$ and $x_j$ are two feature vectors in the input space, and $\gamma$ is a parameter that defines the width of the kernel. The term $||x_i - x_j||^2$ represents squared Euclidean distance between the feature vectors. Additionally, the parameter $\gamma$ is a key parameter in the radial basis function kernel, which essentially controls the scale of the kernel function. A small $\gamma$ value makes the contour of the kernel function broad, which means the influence of a single training example reaches far, including a wide range of other points. Conversely, a large $\gamma$ value makes the contour narrow, implying that the influence of the training examples is limited to a close neighborhood of the point. The radial basis function kernel, through its exponential function, projects the original non-linear observations into a higher-dimensional space without the need to compute the coordinates of the data in that space. This feature, known as the "kernel trick," allows the support vector machine to fit the maximum-margin hyperplane in the transformed feature space. In this new space, the data that is non-linearly separable in the original space could become linearly separable.

In meteorological data, hazardous weather events tend to be rare, which can lead to a serious imbalance in categories and affect the learning effect of the model. To solve this problem, the synthetic minority oversampling technique (SMOTE) can be employed. Synthetic minority oversampling technique balances the class distribution by artificial synthetic enhancement of minority samples. The basic idea is to randomly select a sample from its nearest neighbour for each minority sample, and then randomly generate a new sample between the two.

*3.4. Advantages*
The decision to choose a support vector machine (SVM) over a deep learning approach in predicting hazardous flight weather scenarios can be based on a number of considerations. First, SVMs excel at handling smaller datasets, which is especially important for applications that may not have large amounts of historical meteorological data. Second, SVMs are more economical in computing resource consumption than deep learning models, which makes them effective in resource-constrained environments. In addition, SVM model results are often easier to interpret, which is particularly critical in the field of flight safety, where a clear understanding of the basis for model decisions is required. Explanatory models can help developers and professionals identify and address potential security pitfalls. The working mechanism of SVM—separating different classes by finding the hyperplane of the largest spacing in the data—makes it particularly effective under certain conditions, such as when the data is approximately linearly divisible after proper transformation. In contrast, while deep learning is capable of handling complex nonlinear patterns, it requires large amounts of data to avoid overfitting, and the process of training and debugging models is often more complex and time-consuming. In addition, the black-box nature of deep learning models can pose a barrier in applications that require a high degree of transparency.

## 4. Experiments

*4.1. Experimental Setups*
We utilize the dataset of The NOAA/NWS Storm Prediction Center (SPC), whic provides comprehensive datasets on severe weather events, including tornadoes, thunderstorms, and wind storms, which are essential for forecasting and analyzing severe weather patterns. The SPC offers tools for real-time tracking, mesoanalysis, convective outlooks, and climatology data, facilitating immediate weather response and long-term trend analysis.

Additionally, the SPC supports community outreach and education on weather safety and preparedness. The experimental setup includes collecting historical weather and flight data, normalizing and balancing the data, and selecting relevant features through correlation analysis. Our proposed model utilizes support vector machine with radial basis function kernels, and compares with existing prediction methods including random forests (RF) and long short-term memory networks (LSTMs) were used to adjust hyperparameters through grid search and cross-validation.

### 4.2. Experimental Analysis

Initially, the metric prediction accuracy is a commonly used evaluation metric in machine learning and classification tasks to measure how correct a model's predictions are. Accuracy represents the number of samples that the model correctly classified, while false positives and false negatives represent the number of samples that the model misclassified. Subsequently, the accuracy is intuitive and easy to understand, and is suitable for cases where the class distribution is balanced, but can be misleading when the classes are unbalanced. Figure 1 compares the prediction accuracy results among existing methods.

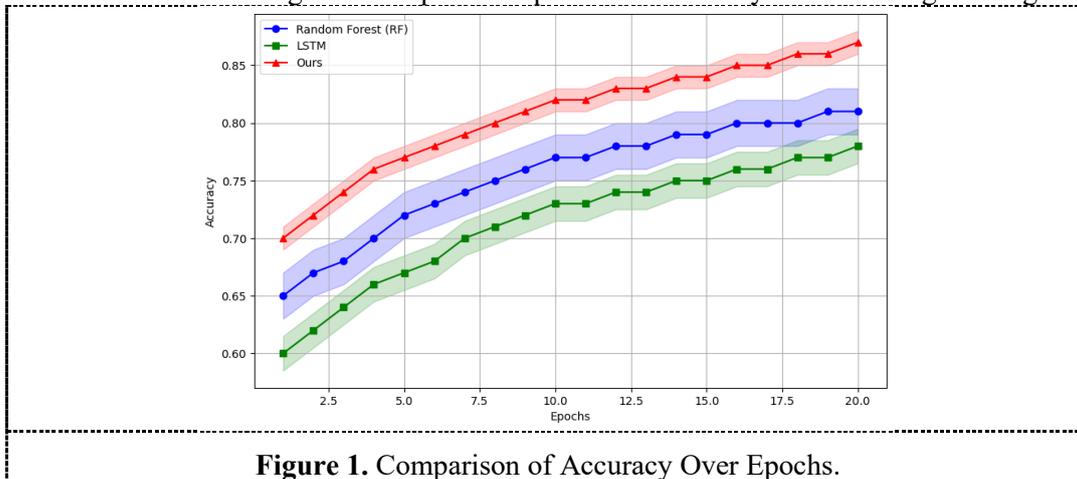

**Figure 1.** Comparison of Accuracy Over Epochs.

Additionally, the ROC-AUC means area under the receiver operating characteristic curve to assess the trade-off between true positive rate and false positive rate. Figure 2 shows ROC-AUC comparison result.

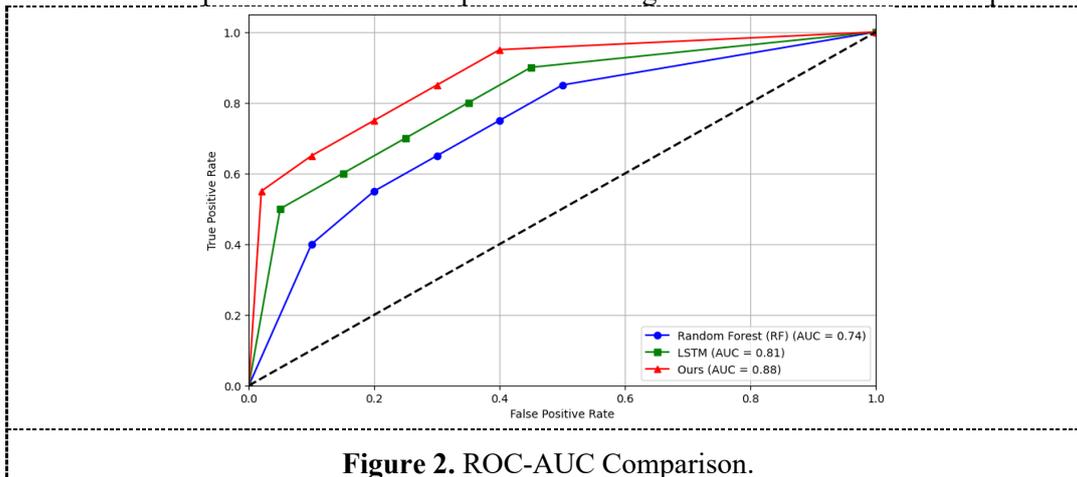

**Figure 2.** ROC-AUC Comparison.

The ROC-AUC comparison chart illustrates the performance of the three methods (Random Forest, LSTM, and ours) in the classification task. The lines of different colors and styles in the diagram represent different methods, and the AUC values for each method are shown in the legend. Our method (red line) exhibits the highest AUC values, indicating the best performance in predicting hazardous flight weather. Additionally, another metric F1 Score means of precision and recall to provide a balanced measure of the model's performance. Following Figure 3 demonstrates the F1 scores comparison results with existing prediction models.

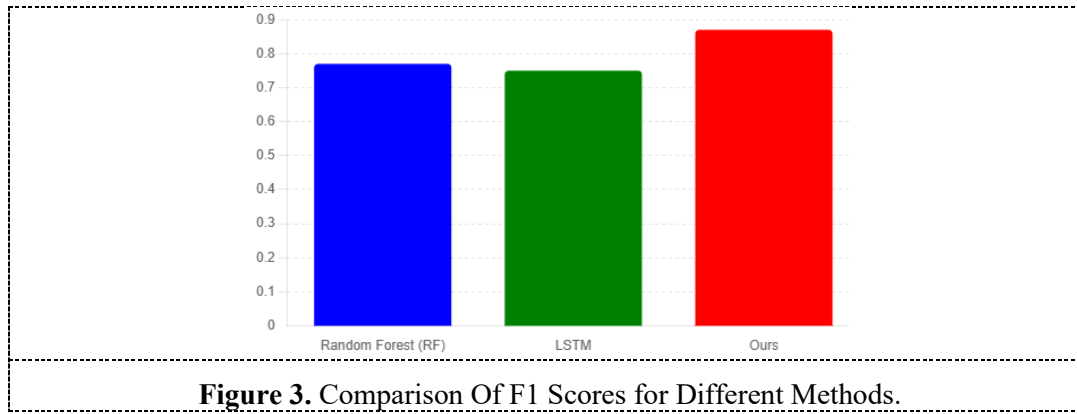
**Figure 3.** Comparison Of F1 Scores for Different Methods.

## 5. Conclusion

In conclusion, we used three machine learning methods: Support Vector Machine (SVM) to predict dangerous flight weather. Through experimental comparison, our method outperformed other methods in multiple evaluation indicators such as accuracy, ROC-AUC and F1 score. Specifically, our method demonstrates higher stability and prediction accuracy during training, effectively improving the early warning capability of dangerous weather. By combining a variety of advanced machine learning technologies and data processing methods, our work provides strong technical support and data guarantee for improving aviation flight safety.